%% file: main.tex
\pdfoutput=1
\documentclass{article}
\usepackage{iclr2027_conference,times}
\input{math_commands.tex}

\usepackage{hyperref}
\usepackage{url}
\usepackage{graphicx}
\usepackage{booktabs}
\usepackage{placeins}
\usepackage{flafter}
\usepackage{float}
\usepackage{needspace}

\title{Is it RL or Data?\\ A Controlled Comparison of GRPO and\\ Rejection-Sampling Fine-Tuning}

\author{Lakshya Chaudhry \\
Purdue University \\
\texttt{lchaudhr@purdue.edu}}

\iclrfinalcopy

\begin{document}
\maketitle
\lhead{Preprint.}

\begin{abstract}
\citet{bauer2026} showed that in low-data reinforcement learning with verifiable rewards, training GRPO on mixed-difficulty prompts gives up to $5\times$ the sample efficiency of easy-only prompts. They did not test whether this benefit is specific to GRPO, or whether another post-training method given the same data would obtain it too. We test this on their procedural counting task with Qwen3-4B-Base and LoRA, comparing GRPO to rejection-sampling fine-tuning (RFT) on the same prompts, verifier and token cap. Six primary arms cross the two methods with three prompt sets (Easy, Mixed, and a Curated subset of prompts the base model sometimes solves, which makes GRPO’s implicit filter explicit). We use three seeds per arm. Hypotheses and falsifiers were registered before training. To rule out a poorly tuned baseline, RFT receives a 27-run hyperparameter sweep, while GRPO uses Bauer et al.'s recipe unchanged. GRPO outperforms RFT at every prompt set: by 10.9 points on held-out test problems with curated prompts, and, in post hoc comparisons with matched rollout budgets, by 8.4 and 9.2 points on the easy and mixed prompts. On test the advantage survives the least favorable treatment of truncated completions; out of distribution, truncation prevents a conclusion. Curating prompts does not help RFT ($-2.7$ points). Mixed-difficulty data helps RFT (+3.7 points), while for GRPO the evidence is inconclusive. A secondary iterated-RFT arm, which resamples from the current policy, recovers 3.2 of the 10.9-point gap (about 30\%). These are controlled measurements on one synthetic task family and one base model with three seeds per arm.
\end{abstract}

\section{Introduction}
\label{sec:intro}

Reinforcement learning with verifiable rewards, most often through GRPO \citep{shao2024deepseekmath}, has become a standard way to improve reasoning in language models, but it also changes which data the model learns from. The \emph{verifier} is a program that checks a completion's final answer against the known correct answer and assigns a score; GRPO uses this score as its reward. Rejection-sampling fine-tuning (RFT; \citealp{yuan2023scaling}) uses the same verifier differently: it samples completions, keeps those the verifier marks correct, and fine-tunes on them. The two methods are hard to compare because GRPO's objective is also a data filter: a prompt whose completions are all right or all wrong gives GRPO no learning signal, so it effectively trains on prompts of intermediate difficulty. Prior work finds that GRPO learns more efficiently from mixed-difficulty data \citep{bauer2026}, and attributes RL's advantage over supervised fine-tuning to this implicit filtering \citep{lu2026why}. If that is right, the advantage should be recoverable by giving a supervised method the same data. We ask: with the same prompts and the same verifier, is GRPO's advantage over other post-training methods, like RFT, unique to the RL objective, or can RFT recover it from the same data? Related studies find that filtered fine-tuning on samples from the current policy performs close to GRPO \citep{xiong2025minimalist,samineni2025rl}; we test whether data sampled once from the base model is enough (\S\ref{sec:related}).

We study this in the low-data setting of \citet{bauer2026}, who train Qwen3-4B with GRPO on a procedural counting task, where mixed-difficulty data gave up to $5\times$ the sample efficiency of easy-only data. The task has one integer answer per problem, so every completion can be verified exactly. We inherit their generator, model family, LoRA configuration and GRPO recipe, and add what the comparison needs: an RFT baseline trained on the same 100-prompt sets, matched rollout budgets (19{,}200 sampled completions per arm on the easy and mixed sets), three seeds per arm, and hypotheses and falsifiers registered before any training run.

Our design crosses three prompt sets (Easy, Mixed, and a Curated subset of Mixed restricted to prompts the base model sometimes solves) with two training signals (RFT and GRPO), giving six arms with three seeds each, plus two single-seed GRPO controls with random and format-only rewards (\S\ref{sec:methodology}, Table~\ref{tab:arms}). Every arm uses the same scoring rule and the same 4{,}352-token cap, and a completion that reaches the cap scores zero. To make sure any GRPO advantage is not an artifact of a poorly tuned baseline, RFT receives a 27-run hyperparameter sweep while GRPO runs a single recipe. A secondary arm, iterated RFT, tests how much of the gap comes from sampling from the current policy.

Our findings are:
\begin{itemize}
  \item \textbf{GRPO outperforms tuned RFT at every prompt set.} On the curated prompts GRPO leads by 10.9\,pp on test. In two comparisons added after the test results were seen, it also leads on the easy and mixed prompts, where the rollout budgets are matched (8.4 and 9.2\,pp). On test the advantage survives the least favorable treatment of truncated completions; out of distribution, truncation is too frequent to draw conclusions.
  \item \textbf{Curating prompts does not help RFT.} RFT trained on the curated prompts scores 2.7\,pp below RFT on the full mixed set, contrary to our pre-registered prediction.
  \item \textbf{Mixed-difficulty data helps RFT; for GRPO the evidence is inconclusive.} Mixed data improves RFT by 3.7\,pp; for GRPO the gain is similar in size but falls short of our criterion.
  \item \textbf{Iterated RFT accounts for about 30\% of the gap.} Three rounds of sampling from the current policy recover 3.2 of GRPO's 10.9\,pp advantage on the curated prompts.
\end{itemize}

These are controlled measurements on one synthetic task family and one base model, trained with LoRA and three seeds per arm; they are not a scaling law for RL with verifiable rewards (\S\ref{sec:limitations}).

\section{Methodology}

\label{sec:methodology}

\subsection{Task and data}
\label{sec:data}

We use the procedural counting task of \citet{bauer2026}. A problem gives an integer range, applies 1--4 filters and 0--3 transformations, and then applies a final operation such as count, sum, or median. The answer is one integer, computed by running the pipeline. We tighten this generator in two ways: every step must change the set, and generation is stratified by range size and step count (Appendix~\ref{app:task}).

\paragraph{Pools.} Training and in-distribution evaluation draw from a pool of 6{,}000 problems with ranges of span 10--200 and 2--5 steps. A separate out-of-distribution set, \texttt{ood\_hard\_200}, uses ranges of span 201--1{,}000 and 6--8 steps, which no training, validation, or selection run ever sees.

\paragraph{Difficulty tiers.} \citet{bauer2026} tier problems by how many frontier models solve them. We tier by the model we train. We sample 8 completions per problem from the base model at $T{=}1.0$ and label a problem \emph{easy} if $\geq 6/8$ are correct, \emph{medium} if 2--5 are, and \emph{hard} if $\leq 1$ is. Unlike an external panel, the trained model's own pass rate measures difficulty
where it matters for learning: whether a group of $G{=}8$ GRPO samples contains both successes and failures depends on the policy's success rate. The same eight samples also define \texttt{train\_curated}, so difficulty and informativeness are measured on one scale.

\paragraph{Data splits.} The splits are disjoint by problem and by pipeline structure:
\begin{itemize}
  \item \texttt{train\_easy\_100}: 100 easy problems.
  \item \texttt{train\_mixed\_100}: 33 easy / 33 medium / 34 hard.
  \item \texttt{train\_curated}: the 73 problems in \texttt{train\_mixed\_100} with between 1 and 7 correct of 8 (26 easy / 33 medium / 14 hard). These are the prompts that would give a GRPO group a learning signal. The set is fixed once and never re-selected.
  \item \texttt{val\_mixed\_100}: 33 / 33 / 34, the only split used for selection.
  \item \texttt{test\_300}: 100 per tier.
\end{itemize}
We also report GSM8K (the first 500 test problems) as a secondary transfer set.

\paragraph{Prompt, scoring, and token cap.} All base-model runs use one plain-text prompt, with no chat template, as is
usual for RL from a base model \citep{deepseek2025r1}: it asks the model to reason step by step and end with a line \texttt{Answer: <integer>}. The reasoning is ordinary text before that line and no \texttt{<think>} tags are used. Scoring checks the number, not the format:
\begin{itemize}
  \item The answer is the last integer on an answer line, else in a \verb|\boxed{}|, else anywhere in the completion.
  \item Correctness is exact match.
  \item The reward for every primary arm is this binary correctness, with no format term.
\end{itemize}

Every run uses a cap of 4{,}352 completion tokens, for both training and evaluation. The cap is derived from base-model completion lengths, with \citeauthor{bauer2026}'s 2{,}048 as a floor (Appendix~\ref{app:cap}). \textbf{A completion that reaches the cap scores zero}, in GRPO rewards, in RFT filtering, and at evaluation.

\FloatBarrier
\subsection{Arms}
\label{sec:arms}

We cross the prompt set (Easy, Mixed, Curated) with the training signal (RFT or GRPO), which gives six arms with three seeds each (Table~\ref{tab:arms}). All arms train a LoRA adapter (rank 64, $\alpha{=}16$, all linear layers) on Qwen3-4B-Base with AdamW, a cosine schedule with 10\% warmup, and gradient clipping at 1.0. Every run is evaluated at its last step.

\begin{table}[t]
\centering\small
\caption{Arms and budgets (means per seed). Rollouts: sampled completions available / used for training. RFT-Mixed has a correct sample for only 95 of its 100 prompts.}
\label{tab:arms}
\begin{tabular}{lccccr}
\toprule
Arm & Seeds & Prompts & Rollouts (avail.\,/\,used) & Train tokens & Opt.\ steps \\
\midrule
RFT-Easy      & 3 & 100 & 19{,}200 / 13{,}834 & 47.2M & 6{,}920 \\
RFT-Mixed     & 3 & 100 & 19{,}200 / 7{,}961  & 15.9M & 1{,}992 \\
RFT-Curated   & 3 & 73  & 14{,}016 / 6{,}653  & 13.4M & 1{,}664 \\
GRPO-Easy     & 3 & 100 & 19{,}200 / 19{,}200 & 12.4M & 300 \\
GRPO-Mixed    & 3 & 100 & 19{,}200 / 19{,}200 & 23.0M & 300 \\
GRPO-Curated  & 3 & 73  & 19{,}200 / 19{,}200 & 20.6M & 300 \\
\midrule
IterRFT-Curated (secondary) & 3 & 73 & 14{,}016 / 7{,}828 & 16.7M & 1{,}963 \\
C1 random reward  & 1 & 100 & 19{,}200 / 19{,}200 & 9.3M  & 300 \\
C2 format reward  & 1 & 100 & 19{,}200 / 19{,}200 & 10.8M & 300 \\
\bottomrule
\end{tabular}
\end{table}

\paragraph{RFT.} Rejection-sampling fine-tuning~\citep{yuan2023scaling} samples completions from a model, keeps those a verifier marks correct, and fine-tunes on them. We sample 192 completions per prompt from the base model once, at $T{=}1.0$, and all RFT arms and seeds share this draw. Each arm keeps the correct completions for its prompts, removes exact duplicates, and fine-tunes on them with a completion-only loss. No other data is mixed in. Because the kept set follows the base model's pass rate, it is dominated by easy prompts: in the RFT Mixed split, only 595 of 7{,}961 kept samples come from hard prompts.

\paragraph{GRPO.} Group Relative Policy Optimization~\citep{shao2024deepseekmath}
is a policy-gradient method that replaces a learned value function with a group baseline: each prompt is sampled several times, and each completion's advantage is its reward normalized by the group's mean and standard deviation. We use the \texttt{GRPOTrainer} of TRL~\citep{vonwerra2020trl} (v1.13.0) for 300 steps. Each
step takes 8 prompts, with 8 completions per prompt, at a learning rate of $5{\times}10^{-5}$. The KL coefficient is 0, and there is one update per sampled batch. The loss is normalized at the token level (\texttt{loss\_type=dapo}; \citealp{yu2025dapo}). Groups whose 8 rewards are all equal contribute no gradient, which is the implicit filtering that \texttt{train\_curated} makes explicit. Capped completions stay in the loss with reward 0. Rollouts come from vLLM~\citep{kwon2023vllm} with the current adapter at every step.

\paragraph{Iterated RFT.} Iterating RFT, where each round samples from the latest policy, is the approach of ReST and ReST$^{EM}$~\citep{gulcehre2023rest,singh2024beyond}. This arm runs three rounds of RFT on the 73 curated prompts, with 64 samples per prompt per round. Round 1 is taken from the base draw, and rounds 2--3 are sampled from the current policy. Each round continues training the previous adapter. The arm refreshes the sampling policy 3 times, against GRPO's 300, and keeps RFT's correct-only objective. It is a secondary arm (\S\ref{sec:eval}).

\paragraph{Controls.} The controls run the GRPO recipe on \texttt{train\_mixed\_100}, one seed each:
\begin{itemize}
  \item \textbf{C1} uses a random Bernoulli(0.5) reward.
  \item \textbf{C2} rewards any uncapped completion that states an explicit final answer, whether or not it is correct.
\end{itemize}

\subsection{Budgets and tuning}
\label{sec:budgets}

\paragraph{Budgets.} Each RFT/GRPO pair trains on the same prompts, and the Easy and Mixed pairs are also matched on rollouts (Table~\ref{tab:arms}). GRPO samples 19{,}200 rollouts in total (300 steps $\times$ 8 prompts per step $\times$ 8 completions per prompt), and RFT draws the same 19{,}200 up front as 192 completions for each of the 100 prompts. On the curated set, RFT has the 14{,}016 samples that belong to its 73 prompts ($73 \times 192$), while GRPO still samples 19{,}200. Our central comparison, the Curated pair, is therefore matched on prompts but not on rollouts. We do not equalize training tokens or optimizer steps: RFT trains for several epochs on correct samples only, while GRPO trains once on every sample, correct or not. That asymmetry is deliberate. It is the mechanism we compare: GRPO also learns from failed samples and from how each sample ranks within its group, while RFT sees only successes.

\paragraph{Tuning.} We tune RFT and not GRPO, so that any advantage for GRPO cannot be explained by a poorly tuned baseline. Each RFT arm gets a $3{\times}3$ sweep over learning rate ($10^{-5}$, $5{\times}10^{-5}$, $10^{-4}$) and epochs (2, 4, 8), 27 runs in total, and we keep the configuration with the best validation accuracy. GRPO runs a single recipe with no search, taken from \citet{bauer2026}: LoRA configuration, learning rate, schedule, 300 steps, and 8 completions per prompt at $T{=}1.0$. We depart from that recipe in two places. Our reward is binary correctness, without their format bonus and step-count penalty, and we use 8 prompts per step (64 completions); \citet{bauer2026} report an effective batch size of 8 without stating whether it counts prompts or completions. Iterated RFT reuses RFT-Curated's configuration. Appendix~\ref{app:hparams} lists all values.

\subsection{Evaluation}
\label{sec:eval}

\paragraph{Metrics.} The primary metric is greedy accuracy on \texttt{test\_300} and \texttt{ood\_hard\_200}, generated with vLLM under the same prompt, scoring rule, and cap for every model. We also report:
\begin{itemize}
  \item mean@8 accuracy at $T{=}1.0$;
  \item pass@$k$ for $k \leq 64$ on the first 100 test problems;
  \item per-tier accuracy;
  \item GSM8K accuracy;
  \item truncation rate for every number.
\end{itemize}
Three reference models are evaluated the same way: Qwen3-4B (non-thinking), Qwen2.5-7B-Instruct, and Llama-3.1-8B-Instruct.

\paragraph{Statistics.} For each arm we report the mean $\pm$ standard deviation over seeds, with every seed shown. Each run also gets a 95\% bootstrap CI over problems. Arm contrasts are paired by seed, with a paired bootstrap CI. A contrast counts as an effect if both of these hold:
\begin{enumerate}
  \item it exceeds twice the pooled seed standard deviation on \texttt{test\_300};
  \item it has the same sign on \texttt{ood\_hard\_200}.
\end{enumerate}
Runs with more than 5\% truncation on \texttt{test\_300} are flagged.

\paragraph{Pre-registration.} The three comparisons (Mixed vs.\ Easy within each method, RFT-Curated vs.\ RFT-Mixed, and GRPO-Curated vs.\ RFT-Curated), our predictions for each, and the success criterion above were committed before any training run (Appendix~\ref{app:prereg}). The iterated-RFT arm, the GRPO--RFT comparisons on the easy and mixed prompts, and two truncation analyses were added after the test results were seen and are marked \emph{secondary} or \emph{post hoc} wherever they appear.

\section{Results}
\label{sec:results}

Table~\ref{tab:main} reports greedy accuracy and truncation for every arm, and Table~\ref{tab:contrasts} reports every paired comparison against the success criterion. We organize the results by comparison, starting with the training signal.

\begin{table}[t]
\centering\small
\caption{Greedy accuracy (\%), mean $\pm$ sd over 3 seeds, with the share of completions cut at the 4{,}352-token cap (trunc.). $^\ddagger$All three seeds exceed 5\% truncation on \texttt{test\_300}. Base, controls and the reference model are single runs.}
\label{tab:main}
\begin{tabular}{lcccc}
\toprule
 & \multicolumn{2}{c}{\texttt{test\_300}} & \multicolumn{2}{c}{\texttt{ood\_hard\_200}} \\
\cmidrule(lr){2-3}\cmidrule(lr){4-5}
Arm & acc. & trunc. & acc. & trunc. \\
\midrule
Base                  & 67.3            & 4.3  & 25.0            & 13.0 \\
RFT-Easy              & 65.9 $\pm$ 1.1  & 2.4  & 25.5 $\pm$ 0.5  & 10.5 \\
RFT-Mixed             & 69.6 $\pm$ 1.0  & 2.0  & 27.3 $\pm$ 1.2  & 12.2 \\
RFT-Curated           & 66.9 $\pm$ 1.1  & 3.1  & 28.5 $\pm$ 1.8  & 14.8 \\
GRPO-Easy             & 74.3 $\pm$ 3.9  & 3.0  & 33.8 $\pm$ 3.1  & 26.2 \\
GRPO-Mixed            & 78.8 $\pm$ 0.5  & 8.7$^\ddagger$ & 30.8 $\pm$ 2.3 & 47.5 \\
GRPO-Curated          & 77.8 $\pm$ 1.7  & 7.6$^\ddagger$ & 31.5 $\pm$ 3.9 & 34.8 \\
\midrule
IterRFT-Curated (secondary) & 70.1 $\pm$ 0.5 & 1.6 & 28.8 $\pm$ 2.4 & 15.0 \\
C1 random reward      & 47.7            & 4.7  & 22.0            & 7.5  \\
C2 format reward      & 63.7            & 1.0  & 27.0            & 10.5 \\
\midrule
Qwen3-4B (reference)  & 74.7            & 4.0  & 45.5            & 22.5 \\
\bottomrule
\end{tabular}
\end{table}

\begin{table}[t]
\centering\small
\caption{Paired comparisons (A $-$ B, pp), mean over seeds with a paired bootstrap
95\% CI. $2\sigma$: twice the pooled seed sd. \checkmark: meets the success criterion
($|\Delta_{\text{test}}| > 2\sigma$ and the same sign on OOD). The easy and mixed GRPO $-$ RFT rows and the iterated-RFT rows are post hoc.}
\label{tab:contrasts}
\begin{tabular}{lcccc}
\toprule
Comparison (A $-$ B) & $\Delta$ \texttt{test\_300} & $2\sigma$ & $\Delta$ \texttt{ood\_hard\_200} & Crit. \\
\midrule
\multicolumn{5}{l}{\emph{Training signal: GRPO $-$ RFT on the same prompts}} \\
GRPO-Curated $-$ RFT-Curated & $+10.9$ [$+6.7$, $+15.1$] & 2.8 & $+3.0$ [$-2.5$, $+8.5$] & \checkmark \\
GRPO-Easy $-$ RFT-Easy       & $+8.4$ [$+4.8$, $+12.1$]  & 5.8 & $+8.3$ [$+3.7$, $+13.0$] & \checkmark \\
GRPO-Mixed $-$ RFT-Mixed     & $+9.2$ [$+4.8$, $+13.7$]  & 1.6 & $+3.5$ [$-2.0$, $+9.0$] & \checkmark \\
\midrule
\multicolumn{5}{l}{\emph{Prompt curation}} \\
RFT-Curated $-$ RFT-Mixed    & $-2.7$ [$-5.2$, $0.0$]    & 2.1 & $+1.2$ [$-2.5$, $+5.3$] & -- \\
GRPO-Curated $-$ GRPO-Mixed  & $-1.0$ [$-3.7$, $+1.7$]   & 2.5 & $+0.7$ [$-2.7$, $+4.0$] & -- \\
\midrule
\multicolumn{5}{l}{\emph{Mixed-difficulty data}} \\
RFT-Mixed $-$ RFT-Easy       & $+3.7$ [$+0.2$, $+7.0$]   & 2.1 & $+1.8$ [$-2.0$, $+5.7$] & \checkmark \\
GRPO-Mixed $-$ GRPO-Easy     & $+4.4$ [$+1.1$, $+7.9$]   & 5.6 & $-3.0$ [$-7.3$, $+1.2$] & -- \\
\midrule
\multicolumn{5}{l}{\emph{Iterated RFT}} \\
IterRFT-Curated $-$ RFT-Curated  & $+3.2$ [$+0.2$, $+6.2$]  & 1.7 & $+0.3$ [$-3.5$, $+3.8$] & \checkmark \\
GRPO-Curated $-$ IterRFT-Curated & $+7.7$ [$+3.7$, $+11.7$] & 2.5 & $+2.7$ [$-2.2$, $+7.3$] & \checkmark \\
\bottomrule
\end{tabular}
\end{table}

\subsection{GRPO outperforms tuned RFT at every prompt set}
\label{sec:res-signal}

When we compare GRPO and RFT on the curated prompts, GRPO is 10.9\,pp more accurate on \texttt{test\_300} (77.8\% vs.\ 66.9\%; Table~\ref{tab:contrasts}). All three seed pairs agree (+12.0, +10.3, +10.3\,pp), and the gap is almost four times the $2\sigma$ threshold. On \texttt{ood\_hard\_200} the gap has the same sign but is smaller, and its interval includes zero (+3.0\,pp, [$-2.5$, $+8.5$]). It meets the success criterion.

Two post hoc comparisons, on the easy and mixed prompts, where the methods are also matched on rollouts, give the same ordering. On the easy prompts GRPO is 8.4\,pp more accurate on test and 8.3\,pp on OOD; neither arm exceeds the truncation flag, and both intervals exclude zero. On the mixed prompts GRPO is 9.2\,pp more accurate on test and 3.5\,pp on OOD.  GRPO's advantage over RFT is therefore of similar size (8--11\,pp on test) whether or not GRPO samples more rollouts than RFT, so the extra rollouts GRPO receives on the curated set cannot account for its advantage there.

\paragraph{RFT changes the model but not its greedy accuracy.} Single-round RFT barely moves greedy test accuracy relative to the base model (Table~\ref{tab:main}): RFT-Easy is 1.4\,pp below it, and the other two arms have intervals that include zero. Relative to Base:
\begin{itemize}
  \item test truncation falls from 4.3\% to 1.7--3.3\%, because RFT learned to stop;
  \item sampled accuracy (mean@8, $T{=}1.0$) rises from 42.5\% to 51.2--54.4\%;
  \item the share of sampled completions with an explicit answer line rises from 77\% to 82--85\%.
\end{itemize}
The format-only control raises mean@8 even further (\S\ref{sec:res-signal}, Controls), so the sampled gain cannot be separated from formatting. On greedy accuracy, which is our primary metric, single-round RFT with the best of nine configurations does not improve on the base model at this budget. GRPO does.

\begin{figure}[t]
\centering
\includegraphics[width=\textwidth]{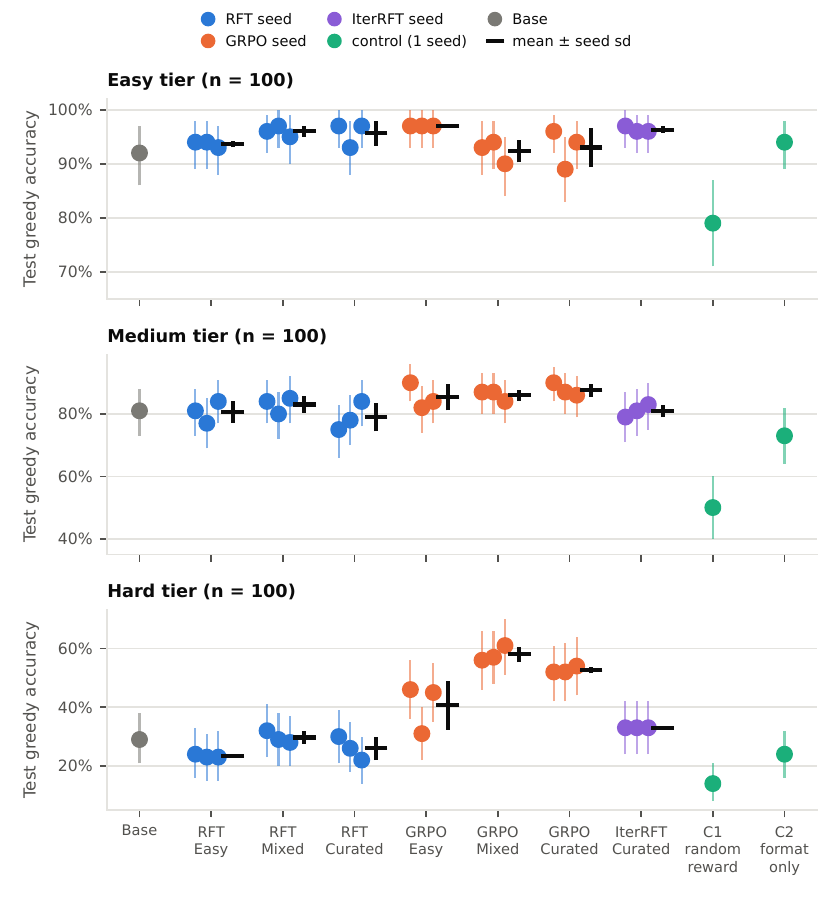}
\caption{Greedy accuracy on \texttt{test\_300} by difficulty tier (100 problems per
tier). Dots are seeds with 95\% bootstrap CIs; black bars are the mean $\pm$ sd over
seeds. Controls and Base are single runs. Each row has its own y-axis range. RFT and
GRPO are close on easy problems and separate on hard ones.}
\label{fig:tier}
\end{figure}

\paragraph{Where the gain lands.} Most of GRPO's gain over Base comes from the hard tier: 85\% of the Mixed arm's gain, 76\% of the Curated arm's, and 56\% of the Easy arm's. On the 100 hard test problems:
\begin{itemize}
  \item Base scores 29.0\%;
  \item the RFT arms score 23.3--29.7\%;
  \item the GRPO arms score 40.7\% (Easy), 58.0\% (Mixed) and 52.7\% (Curated).
\end{itemize}
The hard tier is also where GRPO truncates most (13\% of hard problems for the Mixed and Curated arms), so these GRPO numbers are lower bounds. On the easy tier, where Base already scores 92.0\%, all six primary arms stay between 92\% and 97\% (Figure~\ref{fig:tier}).

\paragraph{Reliability, not coverage.} GRPO makes a single attempt far more likely to succeed: per-sample accuracy (mean@8) rises from 42.5\% for Base to 70.1--77.1\%. It does not enlarge the set of problems the model can ever solve. Given 64 attempts on the first 100 test problems, Base already solves 98.0\%, and the GRPO arms solve 95.7--96.7\%, a difference within the uncertainty of a 100-problem sample. GRPO therefore turns problems the base model solves occasionally into problems it solves reliably, consistent with prior reports that RL with verifiable rewards sharpens the base model's distribution rather than expanding it~\citep{yue2025does} (Appendix~\ref{app:passk}).

\paragraph{Truncation.} GRPO-trained models write longer reasoning traces, so more of their completions hit the 4{,}352-token cap and are scored wrong (1.3--10.3\% per seed on test, against at most 3.3\% for RFT; Tables~\ref{tab:main} and~\ref{tab:seeds}). On test the advantage holds even under the least favorable assumption, crediting every truncated RFT completion as correct and every truncated GRPO completion as wrong (+6.0 to +7.8\,pp across the three prompt sets). On OOD, where 20--55\% of GRPO completions are cut, truncation is large enough to change the differences, so we draw no OOD conclusions from them (Appendix~\ref{app:trunc}).

\paragraph{Controls.} The gain is not reproduced by rewards that ignore correctness. The two controls share GRPO-Mixed's prompts, recipe and seed 1, and differ only in the reward assigned. On test, the correctness reward reaches 78.7\%, the format-only reward 63.7\%, and the random reward 47.7\%, against 67.3\% for Base. On OOD the three score 29.5\%, 27.0\% and 22.0\%, against 25.0\% for Base. Each control is a single seed, so these numbers are descriptive.

The format-only reward does raise mean@8, from 42.5\% to 60.3\%, and pushes the answer-line rate to 99\%. Therefore it is plausible that any sampled-accuracy gain can come from formatting alone.

\subsection{Curating prompts does not help RFT}
\label{sec:res-curation}

Restricting RFT to the prompts with mixed outcomes, the ones that give GRPO a learning signal, does not improve it:
\begin{itemize}
  \item RFT-Curated scores 2.7\,pp \emph{below} RFT-Mixed on test [$-5.2$, $0.0$], with all three seed pairs negative;
  \item on OOD it scores 1.2\,pp above.
\end{itemize}
We had initially predicted that curation would recover most of GRPO's advantage over RFT-Mixed (8.2\,pp). Instead, the recovered share is negative ($-0.32$ as a point estimate), and the prediction is not supported.

The two arms also differ in the learning rate selected on validation ($10^{-5}$ vs.\ $5{\times}10^{-5}$). And a 2.7\,pp gap is about the size of the changes we see when the same run is re-evaluated on a different machine (up to 2.3\,pp on test; \S\ref{sec:limitations}). We conclude that curation does not help RFT, but we do not conclude that it hurts.

\begin{figure}[t]
\centering
\includegraphics[width=\textwidth]{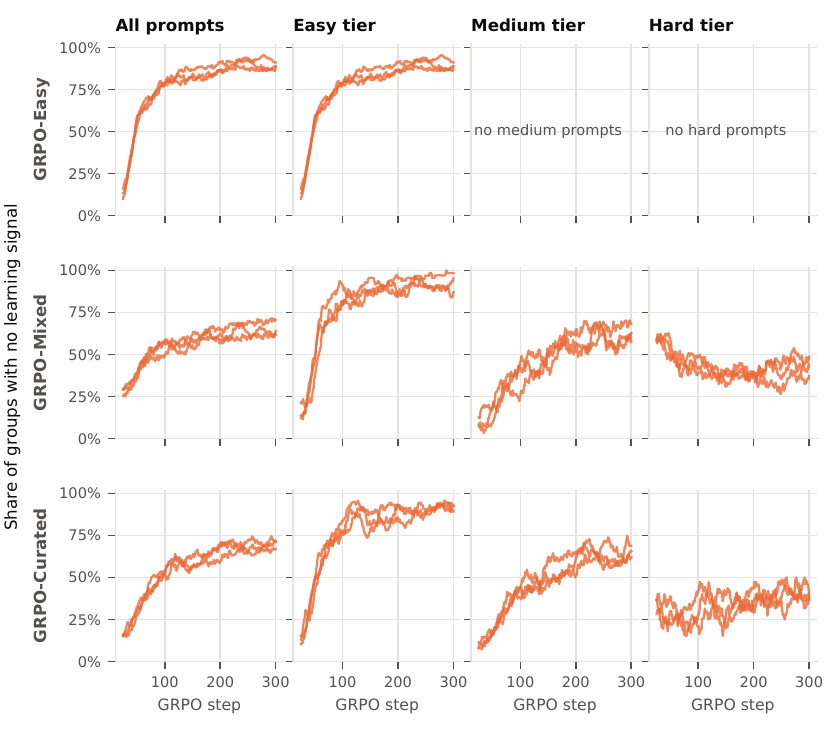}
\caption{Share of GRPO groups whose eight completions all received the same reward, and
so contribute no gradient, over training. One line per seed, 25-step trailing mean.
Columns split the prompts by difficulty tier; the easy prompt set contains only easy
prompts.}
\label{fig:zerostd}
\end{figure}

For GRPO, explicit curation changes nothing detectable: GRPO-Curated is within 1.0\,pp of GRPO-Mixed on test. This matches the implicit-filtering account. GRPO already gets no gradient from groups where every completion agrees, and on the mixed prompts the share of such groups rises from 26--29\% in the first 25 steps to 61--70\% in the last 25 (Figure~\ref{fig:zerostd}). Over training, GRPO filters its own prompts, so filtering them in advance adds little.

\subsection{Mixed-difficulty data helps RFT; for GRPO the evidence is inconclusive}
\label{sec:res-mixed}

\begin{itemize}
  \item \textbf{RFT:} training on the mixed set beats the easy set by 3.7\,pp on test [$+0.2$, $+7.0$], with the same sign on OOD (+1.8\,pp). This meets the criterion.
  \item \textbf{GRPO:} the test difference is similar (+4.4\,pp) but below the threshold, because GRPO-Easy's seeds vary widely (70.0--77.7\%). The OOD sign reverses ($-3.0$\,pp), and every GRPO-Mixed seed is truncation-flagged.
\end{itemize}
The effect \citet{bauer2026} report for RL therefore also appears for SFT on correct samples in our setting. For RL itself, we cannot confirm it, due to the truncation we see for GRPO on OOD.

Two caveats apply to the RFT comparison:
\begin{itemize}
  \item The two arms were selected at different epoch counts (8 vs.\ 4), so the data effect also carries an optimization difference.
  \item The 3.7\,pp gap is only somewhat larger than the cross-machine variation. Because the RFT seeds share one base-model draw, their seed spread, and hence the $2\sigma$ threshold, omits sampling variance.
\end{itemize}

\subsection{Iterated RFT accounts for about 30\% of the gap}
\label{sec:res-iter}
Sampling from the improving policy recovers some of GRPO's advantage, but not most of it.

\paragraph{What iterated RFT tests.} One difference between the methods is where the training data comes from. Single-round RFT learns only from the base model's samples, while GRPO samples from its current policy at every step. Iterated RFT narrows that difference: it runs three rounds of RFT on the curated prompts, and each round samples from the model trained in the previous round. It still learns only from correct
samples.

\paragraph{Results.} Iterated RFT improves on single-round RFT by 3.2\,pp on test [$+0.2$, $+6.2$], with no gain on OOD (+0.3\,pp). GRPO still leads it by 7.7\,pp on test [$+3.7$, $+11.7$].

\paragraph{How much of the gap it explains.} These two steps add up exactly to GRPO's advantage over single-round RFT on the curated prompts ($3.2 + 7.7 = 10.9$\,pp). As a descriptive split, iterated RFT accounts for about 30\% of GRPO's advantage, and about 70\% remains. The remaining 70\% reflects everything GRPO does that iterated RFT does
not: learning from failed samples, comparing samples within a group, refreshing its samples 300 times rather than 3, a larger learning rate, and more rollouts (19{,}200 vs.\ 14{,}016).

Two caveats limit the 30\%. First, it is not purely an effect of fresher samples: iterated RFT also takes more optimizer steps (1{,}963 vs.\ 1{,}664) and trains on more samples (7{,}828 vs.\ 6{,}653). Second, the 3.2\,pp gain is close to the change we see when the same model is re-evaluated on a different machine (up to 2.3\,pp).

\section{Related work}
\label{sec:related}

\paragraph{Fine-tuning on verified samples.} A line of work improves a model by fine-tuning it on those of its own outputs that pass a check. STaR~\citep{zelikman2022star} does this in a loop for rationales that lead to a correct answer. RFT~\citep{yuan2023scaling} collects correct reasoning paths sampled from fine-tuned models and adds them to the fine-tuning data. ReST~\citep{gulcehre2023rest} alternates between sampling from the current policy and improving the policy on those samples with offline RL, and ReST$^{EM}$~\citep{singh2024beyond} repeats a loop of sampling, filtering on binary feedback and fine-tuning a few times. Our RFT arm is a single-round form that samples from the base model and trains on the verified samples alone, and our iterated-RFT arm is the repeated form.

\paragraph{Is RL more than filtered supervised learning?} Several recent studies compare GRPO with simpler methods that train on filtered samples. \citet{xiong2025minimalist} find that RAFT, which keeps only positively rewarded samples, is competitive with GRPO and PPO, and attribute GRPO's main advantage to discarding prompts whose responses are all incorrect. \citet{samineni2025rl} argue that, under the usual modeling assumptions, the GRPO objective reduces to a filtered, iterative form of supervised fine-tuning, and report that this form, using both positive and negative samples, performs comparably to GRPO. \citet{zhu2025surprising} split the RL signal into reinforcing correct samples and penalizing incorrect ones. They find that reinforcing correct samples alone raises pass@1 but lowers pass@$k$ at larger $k$, while penalizing incorrect samples alone often matches GRPO.

Our results differ from the first two: we find a gap of 8--11\,pp between GRPO and RFT. The settings differ in ways that could account for this. Their filtered methods sample from the current policy at every iteration, and \citet{xiong2025minimalist} draw prompts from about 860k problems. Our RFT samples once from the base model and trains for several epochs on at most 100 prompts. Our iterated-RFT arm, which refreshes its samples three times, recovers part of the gap (\S\ref{sec:res-iter}). We did not test an RFT that refreshes at every step, so our results do not contradict theirs. What they show is narrower: data sampled once from the base model was not enough in our setting. Our pass@$k$ results are consistent with \citet{zhu2025surprising} and \citet{yue2025does}: all six primary arms raise per-sample accuracy (mean@8), and none raises pass@64 above the base model (Appendix~\ref{app:passk}).

\paragraph{Difficulty and data selection.} \citet{bauer2026} show that GRPO learns more efficiently from mixed-difficulty prompts in low-data settings. \citet{lu2026why} explain RL's generalization advantage in vision-language models by an implicit filter that favors prompts with mixed outcomes, and show that supervised fine-tuning on a difficulty-filtered training set can surpass GRPO out of distribution. DAPO makes a similar filter an explicit part of RL training, by over-sampling and then dropping prompts whose completions are all correct or all incorrect~\citep{yu2025dapo}. Our \texttt{train\_curated} set matches the medium-difficulty definition of \citet{lu2026why}, a mix of correct and incorrect responses among eight samples from the initial model, and corresponds to their medium-only variant. Their strongest variant also keeps the easy samples, which we did not test as a separate arm. Our setting differs from theirs in two ways. They fine-tune on ground-truth labels for image classification and visual grounding, while RFT fine-tunes on the model's own sampled reasoning. And in RFT the number of training samples per prompt still follows the base model's pass rate. In our setting the filter did not help (\S\ref{sec:res-curation}).

\paragraph{Response length.} Longer outputs under GRPO have been reported before. \citet{liu2025understanding} identify a bias in the GRPO objective that increases response length, especially for incorrect outputs, and \citet{samineni2025rl} argue that the standard formulation indirectly incentivizes longer outputs. Our GRPO arms, which use the token-level loss normalization of DAPO~\citep{yu2025dapo}, also write longer completions than the base model and reach the token cap more often (Appendix~\ref{app:trunc}).

\section{Limitations}
\label{sec:limitations}

This is a controlled comparison on one synthetic task family, not a general result about RL with verifiable rewards.

\paragraph{Task and model.} All training data comes from the procedural counting generator of \citet{bauer2026}, with our two constraints, and we train one base model (Qwen3-4B-Base) with LoRA only. We do not test other task families, model sizes, or full fine-tuning. \citet{bauer2026} study three task families (counting, graph reasoning and spatial reasoning), and we use only counting. With a fixed compute budget (252.7 GPU-hours; Appendix~\ref{app:compute}) we chose three seeds, a tuned baseline and two controls on one task over a second task or model.

\paragraph{Scale of evidence.} Primary arms have three seeds, and the two controls and Base one each. Our success criterion compares a difference to seed variation and checks its sign out of distribution; it is not a formal significance test. The three RFT seeds share one base-model draw, so their spread omits sampling variance. Re-evaluating a model on a different machine changes test accuracy by up to 2.3\,pp. The smaller effects (curation, mixed data for RFT, iterated RFT) are close to that, so we treat them as suggestive.

\paragraph{Budgets and tuning.} The Easy and Mixed pairs match prompts and rollouts; the Curated pair matches prompts only (19{,}200 vs.\ 14{,}016 rollouts). We do not match training tokens or optimizer steps. RFT received a 27-run hyperparameter sweep selected on 100 validation problems, while GRPO used a single recipe from \citet{bauer2026}. This favors RFT, but RFT's selected configurations also differ within its own data comparisons (epochs for mixed data, learning rate for curation).

\paragraph{The RFT baseline.} Three features of our RFT may understate what filtered fine-tuning can do. First, it keeps every correct sample with no per-prompt cap, so its training set follows the base model's pass rate and is dominated by easy prompts (595 of the 7{,}961 samples in RFT-Mixed come from hard prompts), while GRPO samples eight completions for every prompt it draws. A per-prompt cap or tier-balanced sampling might narrow the gap, since GRPO's gain is concentrated on the hard tier (\S\ref{sec:res-signal}). Second, it samples once from the base model, whereas the filtered methods reported to match GRPO resample from the current policy at every iteration \citep{xiong2025minimalist,samineni2025rl}; our iterated arm refreshes only three times. Third, single-round RFT did not improve greedy accuracy over the base model, so the gap between GRPO and RFT is about as large as GRPO's gain over the base model. Our results therefore show that GRPO outperforms this RFT recipe at this budget, not that it outperforms filtered fine-tuning in general. Testing these variants was beyond our compute budget.

\paragraph{What the gap does not isolate.} GRPO's advantage bundles several differences: learning from failed samples, group-relative reward, 300 policy refreshes rather than one or three, a larger learning rate, and more rollouts on the curated set. Iterated RFT separates out part of the refresh effect; we do not separate the rest.

\paragraph{Truncation and scoring.} A completion that reaches the 4{,}352-token cap scores zero, and GRPO reaches it more often. On test, the GRPO advantage survives crediting every truncated completion against it. On OOD, 20--55\% of GRPO completions are
cut, so we draw no OOD conclusions. The extractor falls back to the last integer when no answer line or \verb|\boxed{}| is present; this applies to every arm alike.

\paragraph{Analyses added after seeing test results.} The GRPO--RFT comparisons on the easy and mixed prompts, iterated RFT, and the truncation analyses were defined after the primary test results were seen, and are outside the pre-registered comparisons (Appendix~\ref{app:prereg}).

\section{Conclusion}
\label{sec:conclusion}

We compared GRPO with rejection-sampling fine-tuning (RFT) on the procedural counting task of \citet{bauer2026}, training Qwen3-4B-Base with LoRA. Within each comparison both methods saw the same prompts and the same verifier, and on the easy and mixed prompt sets the same number of rollouts. Our hypotheses, predictions and success criterion were committed before any training run.

GRPO outperformed RFT at every prompt set. On the curated prompts it led by 10.9\,pp on test, and on the easy and mixed prompts, in comparisons added after the test results were seen, it led by 8.4 and 9.2\,pp. On test, this advantage survives the least favorable treatment of truncated completions; out of distribution, truncation prevents a conclusion. Curating prompts did not help RFT, contrary to our pre-registered prediction: RFT on the curated prompts scored 2.7\,pp below RFT on the full mixed set. Mixed-difficulty data helped RFT by 3.7\,pp, while for GRPO the evidence was inconclusive. In a secondary arm, three rounds of iterated RFT recovered 3.2 of the 10.9\,pp gap, about 30\%. None of the data-side changes we gave RFT recovered most of GRPO's advantage.

These results come from one synthetic task family, one base model, LoRA training and three seeds per arm. They measure the two methods under fixed budgets, and are not a general claim about reinforcement learning with verifiable rewards.

\subsection*{Reproducibility statement}
Code, adapters, per-seed metrics, and the pre-registration are available in the project repository at \mbox{\url{https://github.com/LakshyaChaudhry/rl-or-data}}.

\subsection*{AI use statement}
The research question, experimental design, hypotheses and pre-registration are all ours, as are the core components that define the results: answer verification, RFT selection, the SFT loss, the reference GRPO implementation, and the evaluation metrics. An AI coding assistant (Claude) was used to implement the surrounding engineering to our specifications (data generation pipeline, training and evaluation harness, analysis scripts and figures) and to draft and edit the text of this paper from our outlines. We directed and reviewed all of this work, checked every reported number against the logged experiment outputs, and take full responsibility for the paper's content.

\bibliography{references}
\bibliographystyle{iclr2027_conference}

\appendix

\section*{Appendix}
\section{Pre-registration}
\label{app:prereg}

\paragraph{Commitment.} The hypotheses and success criterion below were committed on 2026-09-07 and the predictions on 2026-09-12, before any training run. The first training job started on 2026-09-13. The text is quoted verbatim and has not been edited to match the results. The pre-registration labels the three comparisons H1--H3; in the main text they are the mixed-data, curation, and Curated-pair comparisons.

\paragraph{Hypotheses.}
\begin{quote}\small
\textbf{H1 (data effect):} Mixed-difficulty training data improves held-out performance relative to easy-only data \emph{for both} RFT and GRPO. Test: (RFT-Mixed $-$ RFT-Easy) and (GRPO-Mixed $-$ GRPO-Easy), each vs.\ seed spread. If both are positive, the Bauer et al.\ effect is broader than RLVR.

\textbf{H2 (implicit filtering):} RFT trained on prompts selected by the \emph{same current-policy criterion that makes GRPO groups informative} (1--7 correct of 8) recovers a substantial fraction of GRPO's advantage. Test: (RFT-Curated $-$ RFT-Mixed) relative to (GRPO-Curated $-$ RFT-Mixed). If large, RL's edge is mostly an expensive data-selection mechanism (cf.\ Lu et al.\ 2026, DC-SFT, in VLMs).

\textbf{H3 (on-policy / negative-feedback effect):} GRPO-Curated outperforms RFT-Curated by more than run-to-run variability despite identical prompt and sampling budgets. Residual gain can only come from what RFT lacks: negative samples, relative reward, on-policy generation, repeated policy-dependent exploration.
\end{quote}

\paragraph{Falsifiers.}

\begin{quote}\small
for H1: Falsified by an asymmetry: GRPO-Mixed $>$ GRPO-Easy by more than seed spread, but RFT-Mixed $\approx$ RFT-Easy (or worse). That would mean the mixed-difficulty advantage only exists through the RL objective

for H2: Falsified if RFT-Curated $\approx$ RFT-Mixed -- the same prompt filter that makes GRPO groups informative does nothing for SFT -- while GRPO-Curated stays well above both. Also falsified if RFT-Curated is worse than RFT-Mixed

for H3: Falsified if the difference on test\_300 is within 2$\times$ the pooled seed std, or it flips sign on ood\_hard\_200, or RFT-Curated $\geq$ GRPO-Curated. Any of those means: once data selection is controlled, the RL objective adds nothing you can measure at this budget.
\end{quote}

\paragraph{Success criterion.}
\begin{quote}\small
An arm-vs-arm difference ``counts'' if (a) $|\Delta$ greedy accuracy on test\_300$|$ $>$ 2 $\times$ the pooled seed std of the two arms, and (b) the sign of $\Delta$ is the same on ood\_hard\_200. Report all comparisons regardless.
\end{quote}

\paragraph{Predictions.}
\begin{quote}\small
RFT-all mixed vs easy: RFT-Mixed $>$ RFT-Easy on test\_300 (same direction on ood\_hard\_200), gap $>$ seed noise -- mixed difficulty helps SFT too, so Bauer's effect is not RL-only (H1). GRPO vs RFT-curated: GRPO-Curated $>$ RFT-Curated on test\_300 with the same sign on ood, but the residual is modest; most of GRPO's edge vs RFT-Mixed is recovered by curation (H2), with a leftover from negatives / relative reward / on-policy (H3).
\end{quote}

\paragraph{Logged deviations.}
\begin{enumerate}
  \item \textbf{RFT trains the end-of-sequence token} (2026-09-13). The original RFT recipe did not train the stop token, and validation truncation rose to 36\% during the first sweep. All sweep runs were restarted with the token trained; only validation numbers had been seen.
  \item \textbf{Iterated RFT reuses RFT-Curated's hyperparameters} with no sweep of its own (2026-09-20).
  \item \textbf{Exploratory re-evaluation at a larger cap} (8{,}704 tokens) of Base, RFT-Curated, GRPO-Curated and iterated RFT (2026-09-20), decided after the test results were seen (Appendix~\ref{app:trunc}).
\end{enumerate}

One difference from the registered text is not in this list. H3 as registered states ``identical prompt and sampling budgets''. In the experiment as run, the Curated pair is matched on prompts but not on rollouts: RFT-Curated draws 14{,}016 completions ($73 \times 192$) and GRPO-Curated 19{,}200. The Easy and Mixed pairs are matched on both (\S\ref{sec:budgets}), but those comparisons were added after the test results were seen.

\paragraph{Added after the test results were seen.}
\begin{itemize}
  \item The iterated-RFT arm and its two comparisons, registered on 2026-09-20 before any iterated-RFT sample existed; its prediction was ``small positive --- a few points, maybe not past seed noise'' for iterated RFT over RFT-Curated.
  \item The GRPO--RFT comparisons on the easy and mixed prompts.
  \item The truncation bounds and the larger-cap re-evaluation.
\end{itemize}
The protocol also flags any run whose \texttt{test\_300} truncation exceeds 5\% as not a headline number as it stands. Every seed of GRPO-Mixed and GRPO-Curated exceeds it (5.3--10.3\%); the truncation bounds in Appendix~\ref{app:trunc} are the check we report for those comparisons.

\section{Task generator and splits}
\label{app:task}

\paragraph{Operators.} Following \citet{bauer2026}, a problem is an integer range $[lo, hi]$, then 1--4 filters, then 0--3 transformations, then one final operation. The answer is computed by executing the pipeline in Python and checked in tests against an independent reference implementation.
\begin{itemize}
  \item \textbf{Filters:} even, odd, divisible-by-$N$, not-divisible-by-$N$, below-threshold, above-threshold, digit-sum-equals, contains-digit, prime, perfect-square.
  \item \textbf{Transformations:} add-$k$, multiply-by-$k$, square, absolute value, modulo-$m$, digit sum, reverse digits.
  \item \textbf{Final operations:} count, unique-count, zero-count, sum, product mod $m$, mean (rounded), median, mode, min, max, range, bitwise AND/OR/XOR.
\end{itemize}
Each operator has at least three paraphrases, and connectives depend on position (``First'', ``Then''/``Next'', ``Finally'').

\needspace{4\baselineskip}%
\paragraph{Our constraints.}
\begin{itemize}
  \item Every step must change the set: a pipeline is rejected if a filter or transformation leaves the multiset unchanged, if two consecutive filters are identical, if fewer than three values reach the final operation, if \texttt{mode} or \texttt{unique-count} has no repeated value, or if $|\text{answer}| > 10^9$.
  \item \texttt{positive} and \texttt{negative} filters are removed, since with $lo \geq 1$ they never change the set.
  \item Range spans fall in disjoint bands: S $[10, 50]$, M $[51, 200]$, L $[201, 1000]$.
  \item Generation is stratified, with equal counts per (range band $\times$ step count) cell.
\end{itemize}

\paragraph{Pools and splits.} The main pool has 6{,}000 problems (bands S and M, 2--5 steps, seed 20260901); \texttt{ood\_hard\_200} is generated separately (band L, 6--8 steps, seed 20260902). Splits are disjoint by problem id and by pipeline structure, and were frozen before training.

\begin{table}[!ht]
\centering\small
\caption{Split sizes and tier composition. \texttt{ood\_hard\_200} is tiered after generation but not filtered.}
\label{tab:splits}
\begin{tabular}{lrcc}
\toprule
Split & $n$ & easy / medium / hard & steps \\
\midrule
\texttt{train\_easy\_100}  & 100 & 100 / 0 / 0   & 2--5 \\
\texttt{train\_mixed\_100} & 100 & 33 / 33 / 34  & 2--5 \\
\texttt{train\_curated}    & 73  & 26 / 33 / 14  & 2--5 \\
\texttt{val\_mixed\_100}   & 100 & 33 / 33 / 34  & 2--5 \\
\texttt{test\_300}         & 300 & 100 / 100 / 100 & 2--5 \\
\texttt{ood\_hard\_200}    & 200 & 14 / 47 / 139 & 6--8 \\
\bottomrule
\end{tabular}
\end{table}

\paragraph{Example problems} (from \texttt{test\_300}, verbatim):
\begin{itemize}\small
  \item \emph{Easy (5 steps), answer $-9$:} Start with the inclusive integer range [7, 32]. First, keep only the values that leave no remainder when divided by 7. Next, multiply each remaining number by 3. Next, take every remaining value modulo 7, using the non-negative remainder (0 to 6). Next, subtract 9 from each remaining number. Of these numbers, report the most frequent remaining value, breaking ties by taking the smallest.
  \item \emph{Hard (5 steps), answer 652:} Consider the integers from 3 to 51, inclusive. First, keep only the numbers that are even. Then, keep only the numbers that are divisible by 8. Next, square each remaining number. Then, replace every remaining value x with the digits of $|$x$|$ written in reverse order, with the original sign restored and leading zeros dropped. Of these numbers, report the median of the remaining numbers, taking the lower middle value when the count is even.
\end{itemize}

\section{Answer extraction}
\label{app:extraction}

The extracted answer is the last integer, by position, found in the first of these that is present:
\begin{enumerate}
  \item an \emph{answer line}: a line reading \texttt{Answer: <integer>} once presentation-only decoration is removed (markdown headings, bullets and emphasis; code spans; \texttt{\$\ldots\$}, \verb|\boxed{}|, \verb|\text{}| or angle-bracket wrapping of the integer; an optional ``Final'' and a trailing period or comma);
  \item a \verb|\boxed{<integer>}| anywhere;
  \item the last integer anywhere in the completion.
\end{enumerate}
Digit-group commas are removed. A completion fails extraction only if it contains no integer or reaches the token cap; a truncated completion is never scored from a trailing integer. Taking the last integer means a completion that lists several candidates is scored on the one it ends with.

\paragraph{Why the rule is layered.} An earlier rule required an exact answer line and rejected common formats such as \texttt{**Answer: 36**}. On the 49{,}600-completion tiering run it scored 3{,}326 correct answers as wrong; 23.6\% of the problems it labeled hard (678 of 2{,}843) were answered correctly but in a rejected format. The layered rule reduced extraction failure on that run from 37.1\% to 0.9\%. Where the strict layer and the fallback both produce an answer they agree 98.4\% of the time. The rule was changed after tiering and before any training run, and applied identically to every arm, control and reference model. We call the share of completions answered through layer 1 or 2 the \emph{answer-line rate} (\S\ref{sec:res-signal}).

\section{Token cap}
\label{app:cap}

The cap is $\max\!\left(2048,\ \lceil 1.25 \cdot p_{99,\max} \rceil_{256}\right)$, where $p_{99,\max}$ is the largest per-cell 99th percentile of the length of \emph{correct} base-model completions and $\lceil\cdot\rceil_{256}$ rounds up to a multiple of 256. It was computed once, before the splits existed, from 500 random pool problems $\times$ 8 samples at $T{=}1.0$ (provisional cap 4{,}096; 998 correct completions). The binding cell is M3: $1.25 \times 3{,}477.9 = 4{,}347.4 \rightarrow 4{,}352$. No correct completion in the calibration run reached the provisional cap. Rescoring the same completions under the final extraction rule gives $p_{99,\max} = 3{,}161.7$ and a cap of 4{,}096; we kept 4{,}352 as the more conservative bound.

\begin{table}[!ht]
\centering\small
\caption{Per-cell length statistics of correct base-model completions (tokens) used to set the cap. Cell = range band and total steps.}
\label{tab:cap}
\begin{tabular}{lrrrr}
\toprule
Cell & $n$ correct & median & $p_{99}$ & over 2{,}048 (\%) \\
\midrule
S2 & 139 & 261 & 1{,}183 & 0.0 \\
S3 & 172 & 346 & 1{,}690 & 1.2 \\
S4 & 152 & 463 & 1{,}431 & 0.0 \\
S5 & 133 & 552 & 2{,}114 & 1.5 \\
M2 & 95  & 400 & 2{,}675 & 4.2 \\
M3 & 103 & 392 & 3{,}478 & 2.9 \\
M4 & 122 & 531 & 1{,}543 & 0.8 \\
M5 & 82  & 617 & 2{,}202 & 2.4 \\
\bottomrule
\end{tabular}
\end{table}

\section{Hyperparameters}
\label{app:hparams}

Table~\ref{tab:hparams} lists the training settings and Table~\ref{tab:sweep} the RFT sweep.

\begin{table}[!ht]
\centering\small
\caption{Training settings. Values marked $^\ast$ are library defaults recorded from the resolved configuration.}
\label{tab:hparams}
\begin{tabular}{ll}
\toprule
\multicolumn{2}{l}{\emph{Shared}} \\
Model & \texttt{Qwen/Qwen3-4B-Base}, bf16 \\
LoRA & rank 64, $\alpha{=}16$, dropout 0; q, k, v, o, gate, up, down \\
Optimizer & AdamW, weight decay 0, gradient clip 1.0 \\
Schedule & cosine, 10\% warmup \\
Sampling & $T{=}1.0$, top-$p$ 1.0, cap 4{,}352 completion / 4{,}096 prompt tokens \\
Seeds & 1, 2, 3 (LoRA init, data order, sampling) \\
\midrule
\multicolumn{2}{l}{\emph{RFT}} \\
Draw & 192 samples per prompt from the base model, once (draw seed 1) \\
Acceptance & verified correct; exact duplicates removed; no per-prompt cap \\
Loss & completion-only NLL; end-of-sequence token trained \\
Batch & 16 (micro-batch 4 $\times$ 4 accumulation) \\
Sweep & lr $\{10^{-5}, 5{\times}10^{-5}, 10^{-4}\}$ $\times$ epochs $\{2,4,8\}$, seed 1 \\
Selected & Easy $5{\times}10^{-5}$/8; Mixed $5{\times}10^{-5}$/4; Curated $10^{-5}$/4 \\
\midrule
\multicolumn{2}{l}{\emph{GRPO} (TRL \texttt{GRPOTrainer} 1.13.0, colocated vLLM 0.28.0)} \\
Steps & 300; 8 prompts $\times$ 8 completions per step \\
Learning rate & $5{\times}10^{-5}$ \\
Objective & group-normalized advantage; KL coefficient 0; clip 0.2; \texttt{loss\_type=dapo} \\
Updates & one per generation batch (\texttt{num\_iterations=1}); 64 completions per step \\
Batch & micro-batch 4 $\times$ 16 accumulation \\
Truncated rollouts & reward 0, kept in the loss (\texttt{mask\_truncated\_completions=False}) \\
Optimizer & \texttt{adamw\_torch\_fused}, $\beta = (0.9, 0.999)^\ast$, $\epsilon = 10^{-8\,\ast}$ \\
\midrule
\multicolumn{2}{l}{\emph{Iterated RFT}} \\
Rounds & 3 $\times$ 64 samples per prompt on the 73 curated prompts (14{,}016) \\
Sampling & round 1 from the base draw; rounds 2--3 from the current policy \\
Training & continue previous adapter; fresh optimizer and schedule; lr $10^{-5}$, 4 epochs \\
\midrule
\multicolumn{2}{l}{\emph{Controls} (GRPO recipe on \texttt{train\_mixed\_100}, seed 1)} \\
C1 & reward $\sim$ Bernoulli(0.5), seeded per (seed, problem, step, sample) \\
C2 & reward 1 iff an explicit final answer (layer 1 or 2) and not truncated \\
\bottomrule
\end{tabular}
\end{table}

Other package versions: Python 3.11.16, PyTorch 2.13.0, Transformers 5.16.1, PEFT 0.20.0.

\begin{table}[!ht]
\centering\small
\caption{RFT sweep: \texttt{val\_mixed\_100} greedy accuracy (\%, $n = 100$, seed 1) for each learning rate and epoch count. Bold: selected (ties broken by fewer epochs, then lower learning rate).}
\label{tab:sweep}
\begin{tabular}{l ccc ccc ccc}
\toprule
 & \multicolumn{3}{c}{lr $10^{-5}$} & \multicolumn{3}{c}{lr $5{\times}10^{-5}$} & \multicolumn{3}{c}{lr $10^{-4}$} \\
\cmidrule(lr){2-4}\cmidrule(lr){5-7}\cmidrule(lr){8-10}
Epochs & 2 & 4 & 8 & 2 & 4 & 8 & 2 & 4 & 8 \\
\midrule
RFT-Easy    & 59 & 64 & 61 & 63 & 62 & \textbf{67} & 61 & 64 & 63 \\
RFT-Mixed   & 62 & 62 & 66 & 65 & \textbf{67} & 63 & 62 & 67 & 57 \\
RFT-Curated & 62 & \textbf{66} & 65 & 63 & 64 & 60 & 64 & 64 & 58 \\
\bottomrule
\end{tabular}
\end{table}

\section{Per-seed results}
\label{app:seeds}

Table~\ref{tab:seeds} lists every seed's greedy accuracy and truncation rate.

\begin{table}[!ht]
\centering\small
\caption{Greedy accuracy (\%) by seed, with the share of completions truncated at the cap in parentheses. Mean $\pm$ sd is in Table~\ref{tab:main}.}
\label{tab:seeds}
\begin{tabular}{l ccc ccc}
\toprule
 & \multicolumn{3}{c}{\texttt{test\_300}} & \multicolumn{3}{c}{\texttt{ood\_hard\_200}} \\
\cmidrule(lr){2-4}\cmidrule(lr){5-7}
Arm & seed 1 & seed 2 & seed 3 & seed 1 & seed 2 & seed 3 \\
\midrule
RFT-Easy        & 66.3 (2.3) & 64.7 (3.3) & 66.7 (1.7) & 25.0 (14.0) & 26.0 (9.5) & 25.5 (8.0) \\
RFT-Mixed       & 70.7 (2.7) & 68.7 (1.7) & 69.3 (1.7) & 28.0 (12.0) & 28.0 (11.0) & 26.0 (13.5) \\
RFT-Curated     & 67.3 (2.7) & 65.7 (3.3) & 67.7 (3.3) & 30.0 (14.0) & 29.0 (15.5) & 26.5 (15.0) \\
GRPO-Easy       & 77.7 (4.3) & 70.0 (3.3) & 75.3 (1.3) & 34.5 (38.0) & 30.5 (21.0) & 36.5 (19.5) \\
GRPO-Mixed      & 78.7 (6.7) & 79.3 (10.3) & 78.3 (9.0) & 29.5 (46.5) & 29.5 (54.5) & 33.5 (41.5) \\
GRPO-Curated    & 79.3 (5.3) & 76.0 (9.3) & 78.0 (8.0) & 36.0 (19.5) & 29.0 (45.0) & 29.5 (40.0) \\
IterRFT-Curated & 69.7 (1.3) & 70.0 (1.7) & 70.7 (1.7) & 28.0 (14.5) & 27.0 (16.0) & 31.5 (14.5) \\
\bottomrule
\end{tabular}
\end{table}

\section{Truncation}
\label{app:trunc}

\begin{figure}[!ht]
\centering
\includegraphics[width=\textwidth]{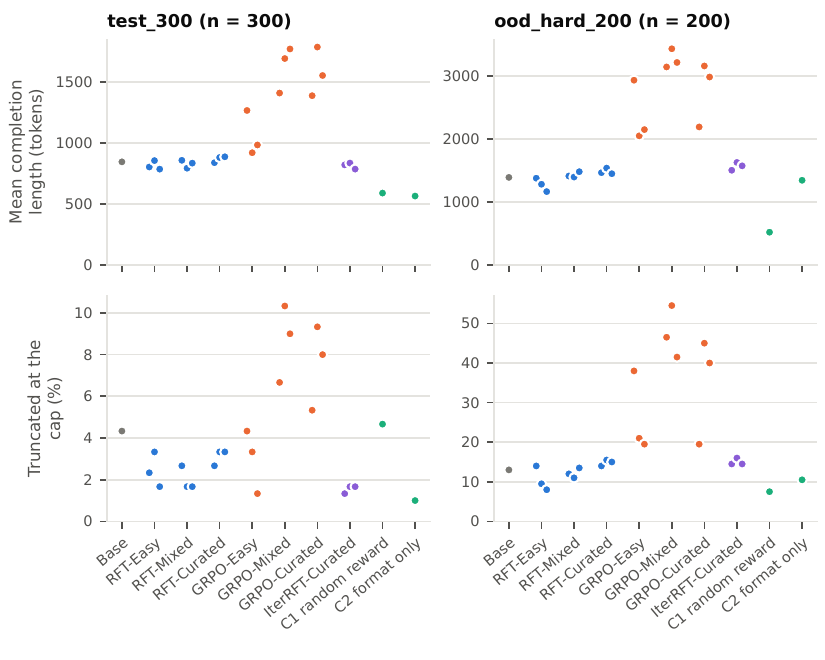}
\caption{Mean completion length (top) and share of completions truncated at the 4{,}352-token cap (bottom) under greedy decoding. Each dot is one seed (three per arm; one for Base, C1 and C2). Colors: gray Base, blue RFT, orange GRPO, purple iterated RFT, green controls. GRPO-Mixed and GRPO-Curated write longer completions and are truncated more often than RFT, especially on the out-of-distribution set.}
\label{fig:evallength}
\end{figure}

\paragraph{Bounds.} A truncated completion is scored wrong, and inspection of the greedy outputs found that truncated completions contain no final answer before the cut. To bound how much truncation could move each comparison, we rescore each run at two extremes: every truncated completion wrong (as reported) or every truncated completion right. The lower bound of a difference A $-$ B credits B's truncated completions and not A's; the upper bound does the reverse. These bounds were defined after the test results were seen.

\begin{table}[!ht]
\centering\small
\caption{Truncation bounds on \texttt{test\_300} (pp): the observed difference and its range when truncated completions are counted against A (lower) or against B (upper). On \texttt{ood\_hard\_200} every interval includes zero.}
\label{tab:bounds}
\begin{tabular}{lccc}
\toprule
Comparison (A $-$ B) & observed & lower & upper \\
\midrule
GRPO-Curated $-$ RFT-Curated & $+10.9$ & $+7.8$ & $+18.4$ \\
GRPO-Easy $-$ RFT-Easy       & $+8.4$  & $+6.0$ & $+11.4$ \\
GRPO-Mixed $-$ RFT-Mixed     & $+9.2$  & $+7.2$ & $+17.9$ \\
RFT-Curated $-$ RFT-Mixed    & $-2.7$  & $-4.7$ & $+0.4$ \\
RFT-Mixed $-$ RFT-Easy       & $+3.7$  & $+1.2$ & $+5.7$ \\
GRPO-Mixed $-$ GRPO-Easy     & $+4.4$  & $+1.4$ & $+13.1$ \\
IterRFT-Curated $-$ RFT-Curated & $+3.2$ & $+0.1$ & $+4.8$ \\
GRPO-Curated $-$ IterRFT-Curated & $+7.7$ & $+6.1$ & $+15.2$ \\
\bottomrule
\end{tabular}
\end{table}

\paragraph{Larger cap (exploratory).} We re-generated greedy completions at twice the cap (8{,}704 tokens) for Base, RFT-Curated, iterated RFT and GRPO-Curated. The models were still trained at 4{,}352. At the larger cap GRPO-Curated leads RFT-Curated by 15.0\,pp on test [$+11.1$, $+19.1$] and 11.5\,pp on OOD [$+6.0$, $+17.0$], and leads iterated RFT by 11.9 and 10.7\,pp. GRPO-Curated is still cut on 11--31\% of OOD problems at the larger cap.

\paragraph{Reproducibility across machines.} Greedy decoding with vLLM was byte-identical when a model was re-evaluated on the same machine (1{,}396 of 1{,}396 completions), but not across machines: 6--10\% of problems changed outcome, moving accuracy by up to 2.3\,pp on test and 5.0\,pp on OOD. Runs were evaluated on several machines, so this variation is not contained in the seed spread. It is small next to the GRPO--RFT differences but comparable to the curation, mixed-data (RFT) and iterated-RFT effects.

\section{pass@\texorpdfstring{$k$}{k} and sampled accuracy}
\label{app:passk}

\begin{figure}[!ht]
\centering
\includegraphics[width=0.8\textwidth]{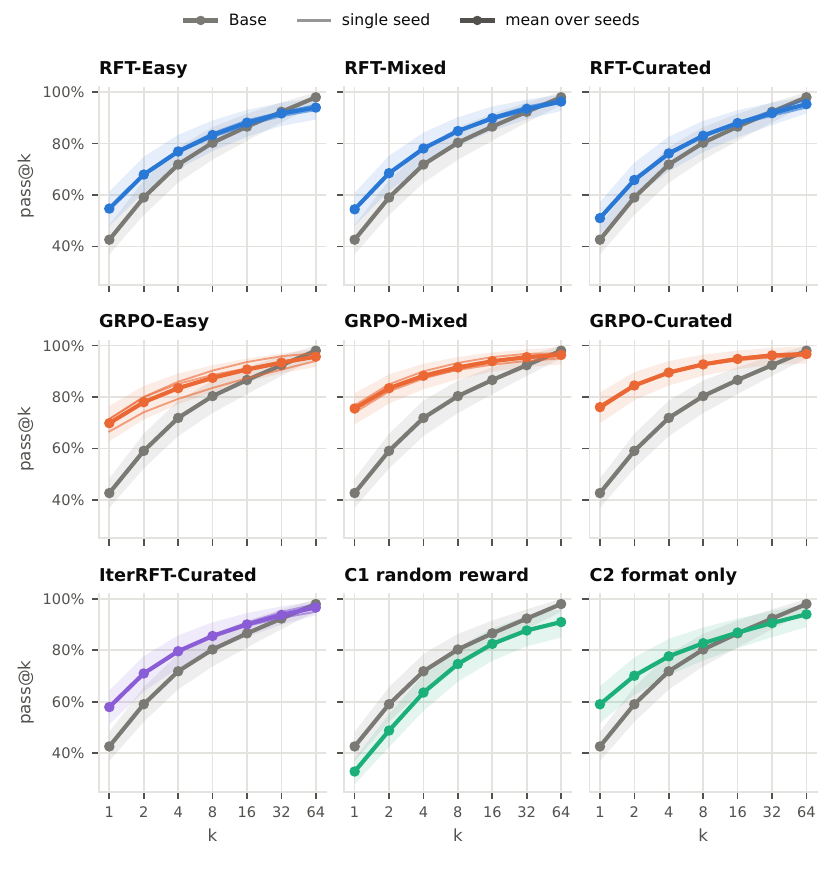}
\caption{pass@$k$ on the first 100 \texttt{test\_300} problems, from 64 samples per problem at $T{=}1.0$ (unbiased estimator). Thin lines are seeds; thick lines are the mean over seeds; shaded bands are 95\% bootstrap CIs over the 100 problems, for Base (gray) as well. Arms are colored: RFT blue, GRPO orange, iterated RFT purple, controls green. C1 and C2 are single seeds (one curve, no seed lines). GRPO lifts pass@$k$ at small $k$ and meets Base, within the bands, at $k = 64$.}
\label{fig:passk}
\end{figure}

Figure~\ref{fig:passk} shows pass@$k$ for every arm, and Table~\ref{tab:sampled} reports sampled accuracy.

\begin{table}[!ht]
\centering\small
\caption{Sampled accuracy (\%): mean@8 at $T{=}1.0$ on \texttt{test\_300} and \texttt{ood\_hard\_200}, and pass@8 / pass@64 on the first 100 test problems. Mean over seeds; controls and Base are single runs.}
\label{tab:sampled}
\begin{tabular}{lcccc}
\toprule
Arm & test mean@8 & OOD mean@8 & pass@8 & pass@64 \\
\midrule
Base            & 42.5 & 17.3 & 80.3 & 98.0 \\
RFT-Easy        & 53.5 & 24.3 & 83.3 & 94.0 \\
RFT-Mixed       & 54.4 & 23.5 & 84.9 & 96.3 \\
RFT-Curated     & 51.2 & 21.6 & 83.0 & 95.3 \\
GRPO-Easy       & 70.1 & 33.0 & 87.4 & 95.7 \\
GRPO-Mixed      & 77.1 & 31.1 & 91.5 & 96.3 \\
GRPO-Curated    & 76.3 & 34.8 & 92.7 & 96.7 \\
IterRFT-Curated & 58.5 & 25.0 & 85.6 & 96.7 \\
C1 random reward & 33.6 & 14.0 & 74.7 & 91.0 \\
C2 format reward & 60.3 & 26.4 & 82.7 & 94.0 \\
\bottomrule
\end{tabular}
\end{table}

\section{Additional results}
\label{app:extra}

\paragraph{GSM8K transfer.} Greedy accuracy on the first 500 GSM8K test problems, same prompt and cap: Base 88.0\%; RFT-Easy 91.9\%, RFT-Mixed 90.9\%, RFT-Curated 91.5\%; GRPO-Easy 93.3\%, GRPO-Mixed 92.7\%, GRPO-Curated 92.9\%; iterated RFT 92.1\%; C1 85.0\%; C2 93.2\%. Truncation is at most 1.0\% for every run. We computed no comparisons on this set; reference models were not evaluated on it.

\paragraph{Reference models.} Greedy \texttt{test\_300} / \texttt{ood\_hard\_200} accuracy, same prompt inside each model's chat template: Qwen3-4B (non-thinking) 74.7 / 45.5\%; Qwen2.5-7B-Instruct 65.3 / 28.0\%; Llama-3.1-8B-Instruct 37.3 / 9.0\% (11.3\% of its test completions truncated).

\section{Compute}
\label{app:compute}

All runs used one NVIDIA H100 PCIe 80\,GB GPU per job. Table~\ref{tab:compute} gives metered GPU-hours per arm, including the RFT sweep and all evaluations. Arms and controls total 212.1 GPU-hours; with base and reference evaluations, tiering, the shared RFT draw, the exploratory re-evaluation, and discarded runs (RFT before the end-of-sequence fix, and a failed GRPO attempt), the metered total is 252.7 GPU-hours.

\begin{table}[!ht]
\centering\small
\caption{Metered GPU-hours (H100) per arm. ``Sweep'' is the eight non-selected RFT configurations.}
\label{tab:compute}
\begin{tabular}{lrrrr}
\toprule
Arm & training (selected) & sweep & evaluation & total \\
\midrule
RFT-Easy        & 21.3 & 30.1 & 1.8 & 53.1 \\
RFT-Mixed       & 7.0  & 22.1 & 1.9 & 31.0 \\
RFT-Curated     & 5.9  & 18.5 & 1.9 & 26.3 \\
GRPO-Easy       & 15.5 & --   & 3.0 & 18.5 \\
GRPO-Mixed      & 27.1 & --   & 5.0 & 32.0 \\
GRPO-Curated    & 25.1 & --   & 4.4 & 29.4 \\
IterRFT-Curated & 8.0  & --   & 1.9 & 9.8 \\
C1 random reward & 5.9 & --   & 0.5 & 6.4 \\
C2 format reward & 5.0 & --   & 0.5 & 5.5 \\
\midrule
Total           &      &      &     & 212.1 \\
\bottomrule
\end{tabular}
\end{table}

\end{document}

%% file: math_commands.tex
\usepackage{amsmath,amsfonts,bm}

\def\eqref#1{equation~\ref{#1}}

\def\1{\bm{1}}

\DeclareMathAlphabet{\mathsfit}{\encodingdefault}{\sfdefault}{m}{sl}
\SetMathAlphabet{\mathsfit}{bold}{\encodingdefault}{\sfdefault}{bx}{n}

